# Automated Spelling Correction for Clinical Text Mining in Russian

Ksenia BALABAEVA[a,1], Anastasia FUNKNER[a] and Sergey KOVALCHUK[a]

[a]*ITMO University, Saint Petersburg, Russia*

**Abstract.** The main goal of this paper is to develop a spell checker module for clinical text in Russian. The described approach combines string distance measure algorithms with technics of machine learning embedding methods. Our overall precision is 0.86, lexical precision - 0.975 and error precision is 0.74. We develop spell checker as a part of medical text mining tool regarding the problems of misspelling, negation, experiencer and temporality detection.

## 1. Introduction

Predictive modeling in medicine and healthcare is developing rapidly, and the quality of modeling results crucially depend on the quality of the input data. In these domains, a significant proportion of the information is presented with texts in natural language. Feature extraction from medical texts is challenging due to the following problems: misprints in medical terms, negation, temporality, and a person (experiencer) that have experienced an event (it could be a patient or his or her relative. In the present study, our goal is to develop module of spelling correction to more accurately process and extract knowledge from clinical records in natural language.

## 2. Related works

Processing of clinical texts in natural language is a well-studied problem, mainly for the English language. There are many tools for labeling text, extracting entities, disease's cases, temporal and negation detection (UMLS, UIMA, IBM Watson, Apache Ruta, etc.), but their require collection of language corpus [1–4]. In some other countries, scientists develop their own corpus-free machine-learning tools or tools that can solve problems which are very specific to their

---

language [5, 6, 7]. To our knowledge, for the Russian language, a corpus of medical texts has not yet been compiled. There is a small case based on 120 records with labeled diseases and their attributes (complications, severity, treatment, etc.) [8,9]. Machine learning and artificial intelligence are increasingly used in medicine and healthcare. As a result, new methods and models are developed in NLP to obtain additional features from texts [2,4].

To extract knowledge from medical texts it is necessary to cope with many challenges: expansion of acronyms and abbreviations, interpretation of domain specific words, handling spelling mistakes, temporal tagging, negation detection [10,11], etc. There are many solutions for above mentioned problems regarding English language (UMLS, UIMA, OHNLP, etc.). However, NLP instruments for Russian are very limited and there is no generally accepted system to process clinical texts in Russian.

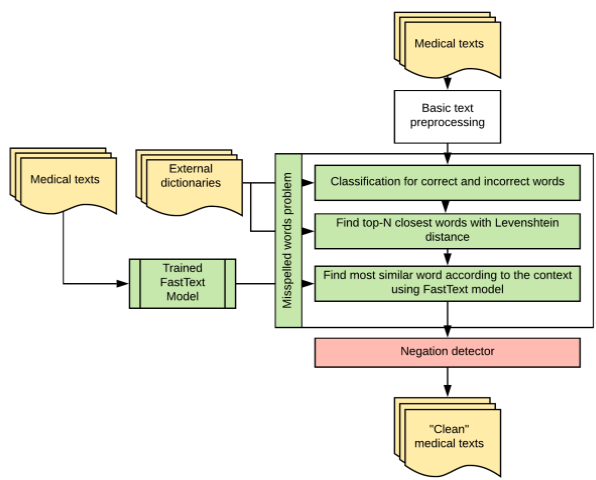

**Figure 1**. Modules of spelling correction and negation detection. Green blocks indicate methods described in this paper. Pink blocks are developed and implemented by colleagues.

We are developing corpus-independent modules with a spellchecker and a negation detector trained on Russian medical texts. Figure 1 shows the scheme of language processing application. And it also can be used for similar languages. In this research, we used a set of 3434 electronic health records (EHRs) of ACS patients who admitted to Almazov National Medical Research Centre (Almazov Centre) during 2010-2015. EHRs includes disease and life anamnesis, doctor's reports, discharge report, etc. in natural language. Disease anamnesis are the most unstructured records and we used them to demonstrate our approach.

In order to correct misspelled words, we need to solve three main tasks: find incorrect words in text, define the type of mistake and correct the word (Figure 1). To identify misspelled words, we use external dictionaries of russian words, medical terms, drugs, abbreviations, etc. We also broaden the pool with specific non-dictionary words of a high frequency from clinical records. Our module works with the following types of mistakes.

**Table 1.** Types of misspelled words.

| Type of mistake | Incorrect text | Corrected text |
|---|---|---|
| Word reduction | anem | anemia |
| Missing/ misspeled symbols | anmia | anemia |
| Multiple words separation | hemolyticanemia | hemolytic anemia |

In current realization we can automatically identify only multiple words mistake type by dividing the word sequence into all possible subsequences and evaluating if each part is an existing word and their probability to stand next to each other in a sentence is high. The basic idea of correction is to replace the incorrect word with the most similar correct word from the dict. Our approach combines machine learning technics for the word-to-vector representation with more traditional approaches of string distance calculations. To measure the performance of spell checker, we use lexical precision (which is treated as a measure of incorrect words identification), error precision and overall linguistic performance.

## 3. Method

The classical approaches to measure words similarity is Levenshtein distance or edit distance and its extensions (Damerau-Levenshtein-DL) [13]. It is often used for spell correction [15]. It calculates words similarity according to the minimal number of basic operations required to get the target word from the other. However, this approach doesn't consider the context provided by the sentence.

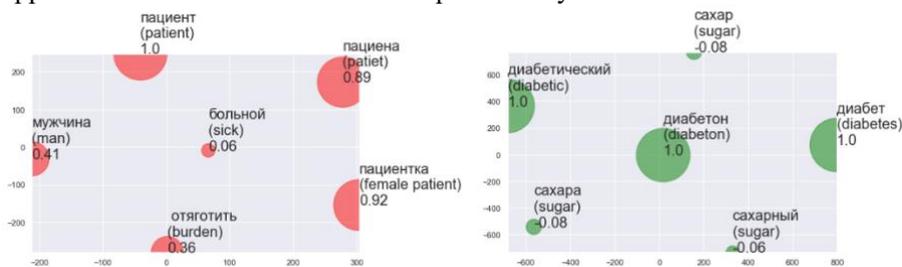

**Figure 2.** Most similar words for word 'patient' based on FastText vector representation (on the right). And Most similar words for the word 'diabetes' based on Word2Vec vector representation (on the left)

In order to map words into numerical vectors we use the following word-embeddings: Word2Vec [12] and also FastText [14] on SG and CBOW. We chose Word2Vec and FastText, since they are widely and efficiently used in different NLP problems, including spell correction, and leverage context of the target word. In simple words both methods build a vector representation using a hidden layer of the Neural Network. The difference between models is in token representation: FastText treats each word as a vector of n-grams, whether W2V uses the whole word. Using such representation, we can easily calculate distance between words with cosine similarity. The example of finding close words is on the Figure 2.

## 4. Results and discussion

The methods were evaluated on hold-out sample of randomly sampled 200 correct medical words and 200 incorrect medical terms. All results are performed in Table 2. According to the results, we may conclude that both string distance-based algorithms and vector-based methods perform poorly when used separately. Lev. distance and DL distance are lacking context information and the 'meaning' of tokens. However, embedding-based approach may confuse correct word for the synonym or other related words. That is why our final model combines these approaches in the following way.

**Table 2**. Results of spell checker experiments

| Method | Lexical Precision | Error Precision | Overall Precision |
|---|---|---|---|
| Levenshtein dist. | 0.975 | 0.52 | 0.7475 |
| Damerau–Levenshtein (DL) dist | 0.975 | 0.545 | 0.76 |
| Cos. dist. on Word2vec | 0.975 | 0.42 | 0.6975 |
| Cos. dist. on FastText CBOW | 0.975 | 0.375 | 0.675 |
| Cos. dist. on FastText SG | 0.975 | 0.405 | 0.69 |
| FastText + DL dist | 0.975 | 0.745 | 0.86 |
| Mean(CBOW, SG, W2V)+DL Dist | 0.975 | 0.745 | 0.85 |

At the first stage, we get top-n similar words according to our word-model trained on anamnesis text, then we select m words from the dictionary and top-n words, by minimal DL distance. We also tried different ensembling approaches (voting, bagging, etc.), however it outperformed the others. The error analysis allows us to suggest, that, in general, Levenshtein distance and DL distance when used separately perform better for misprints and the words with missing letters. However, vector-based methods are better at reduced word spelling (anem, anemia). That is why it could be more precise to use independent approaches for each mistake type in future works.

## 5. Conclusion

We may conclude that in terms of incorrect words detection our approach works steadily and is quite precise 0.975 lexical precision. In terms of error precision it is 0.745, which is high for texts with specialized language, but still there is a room for advancement. The output data can be used for more accurate knowledge extraction from medical texts for the further processing and modelling. In order to improve our approach concerning the problem of spell checking, we are going to expand external dictionaries for more medicine names, rare medical terms, etc. Also, we want to build ML classifier to separately work with different types of mistakes (Table 1). And our main goal in the future is to expend module's functionality towards negation, experiencer and temporality detection and make a tool for clinical text mining in Russian.

**Acknowledgements**

This research is financially supported by The Russian Science Foundation, Agreement #19-11-00326.